# Parametric Leaky Tanh: A New Hybrid Activation Function for Deep Learning

**Stamatis Mastromichalakis[1]**

[1]London South Bank University / IST College,
Pireos 72, GR-18346, Moschato, Athens, Greece
Email: stamatis@tmnetworks.gr

**ABSTRACT**

*Activation functions (AFs) are crucial components of deep neural networks (DNNs), having a significant impact on their performance. An activation function in a DNN is typically a smooth, nonlinear function that transforms an input signal into an output signal for the subsequent layer. In this paper, we propose the Parametric Leaky Tanh (PLTanh), a novel hybrid activation function designed to combine the strengths of both the Tanh and Leaky ReLU (LReLU) activation functions. PLTanh is differentiable at all points and addresses the 'dying ReLU' problem by ensuring a non-zero gradient for negative inputs, consistent with the behavior of LReLU. By integrating the unique advantages of these two diverse activation functions, PLTanh facilitates the learning of more intricate nonlinear relationships within the network. This paper presents an empirical evaluation of PLTanh against established activation functions, namely ReLU, LReLU, and ALReLU utilizing five diverse datasets.*

## 1. INTRODUCTION

Activation functions (AFs) are instrumental in shaping the performance of Deep Neural Networks (DNNs), responsible for transforming input signals into output signals for subsequent network layers. In this study, we propose a novel activation function, Parametric Leaky Tanh (PLTanh) that harnesses the benefits of both Tanh and LReLU to enhance DNN performance.

The Tanh activation function is a smooth, differentiable function that maps real numbers to the interval [-1, 1]. Its bounded nature and symmetric properties around the origin make it a fitting choice for applications that require resilience to outliers or expect output centered around zero.

In contrast, the LReLU activation function, favored in DNNs for its computational efficiency, introduces a slight gradient for negative input values, preventing the 'dying ReLU' problem commonly encountered with traditional ReLU. This ensures that all negative inputs contribute to the learning process.

By merging the properties of Tanh and LReLU, we formulate a new hybrid activation function capable of learning more complex nonlinear relationships. Such a combination can pave the way for models that establish more sophisticated relationships between input and output data, contributing to more robust performance and better generalization.

While Tanh and LReLU are widely employed in DNNs, they each come with their own set of strengths and weaknesses. Our proposed combination seeks to significantly enhance the performance of a DNN by capitalizing on their individual strengths and mitigating their weaknesses. The combined benefits of centering from Tanh and non-zero gradients for negative inputs from LReLU can lead to improved DNN performance.

Despite significant advancements in the development of activation functions, such as the introduction of QReLU/m-QReLU (Parisi et al., 2020a) and ALReLU (Mastromichalakis, 2020), SigmoReLU (Mastromichalakis, 2021), traditional activation functions like the Sigmoid and Tanh are still plagued by the well-known issue of the vanishing gradient problem. Traditional ReLU offers more accuracy and scalability for DNNs but is susceptible to the 'dying ReLU' problem. Several variants of ReLU, such as the Leaky ReLU (LReLU), Parametric ReLU (PReLU), Randomised ReLU (RReLU), and Concatenated ReLU (CReLU) were developed to address these challenges. For instance, LReLU (Maas et al., 2013) provides a small negative slope for negative inputs, leading to minor improvements in classification performance compared to the original ReLU. However, these AFs often encounter issues of robustness in classification tasks of varying complexity, such as slow or non-convergence (Vert and Vert, 2006) and frequently fall into local minima (Parisi et al., 2020b).

In this study, we introduce a novel variant of the tanh AF to alleviate common vanishing gradient and 'dying ReLU' issues. Based on numerical evaluations, our method offers substantial improvements in training and classification procedures compared with ReLU, LReLU, and ALReLU across five distinct datasets. Evaluation metrics such as accuracy, AUC, recall, precision, and F1-score were computed to assess the performance of our proposed technique and provide a reliable, objective basis for comparison.

The rest of this paper is structured as follows: Section 2 provides the main contribution of this study, detailing the implementation of PLTanh in Keras. Section 3 presents experimental results and an evaluation of the training accuracy, comparing PLTanh with other established AFs in the field. Finally, Section 4 concludes with a discussion and summary of our findings..

## 2. METHODS AND ALGORITHMS

### 2.1 Datasets and Models Used for Training

The experiments in this study utilized various datasets encompassing image, text, and tabular data classifications. The specific datasets employed were:
- MNIST Dataset
- Fashion MNIST Dataset

- TensorFlow Flowers Dataset
- CIFAR-10 Dataset
- Histopathologic Cancer Detection Dataset used in the 2019 Kaggle competition (https://www.kaggle.com/c/histopathologic-cancer-detection/data)

For the training of the MNIST and Fashion MNIST datasets, a deep Convolutional Neural Network (CNN) model was used, with the following layers:

- A convolutional layer consisting of 32 filters, each with a kernel size of 3x3. The corresponding activation function (AF) was applied after this layer.
- A Max Pooling 2D layer for downsampling the input.
- A Flatten layer to transform the 2D matrix data to a 1D vector.
- A final Dense layer with softmax activation for outputting probabilities for the classes.

For the TensorFlow Flowers dataset, a deeper Convolutional Neural Network (CNN) model was utilized. The architecture is as follows:

- The first layer is a Conv2D layer equipped with 32 filters of size 3x3, utilizing the corresponding activation function under test in each respective case. The layer also processes an input shape of (32, 32, 3).
- A Max Pooling 2D layer is then used for downsampling the input representation, followed by a Dropout layer with a rate of 0.25 to reduce overfitting.
- A second Conv2D layer with 64 filters of size 3x3 is then added, again using the corresponding AF, followed by another Max Pooling 2D layer and Dropout layer (with the same rate of 0.25).
- A third Conv2D layer is then applied, this time with 128 filters of size 3x3 and the corresponding AF, followed by a Dropout layer with a rate of 0.4.
- The data is then flattened from a 2D matrix to a 1D vector using a Flatten layer.
- This is followed by a Dense layer with 128 units and the corresponding AF, and another Dropout layer with a rate of 0.3.
- Finally, the output layer is a Dense layer with 5 units (representing the number of classes in the Flowers dataset) and a softmax activation function for outputting the probability distribution across the classes.

For CIFAR-10 Dataset The following CNN was used:

- The first layer of the model is a Conv2D layer with 32 filters of size 3x3, using the corresponding AF under test. This layer applies 'same' padding and accepts an input shape of (32, 32, 3). This is followed by a Batch Normalization layer.
- This is followed by another Conv2D layer with 32 filters of size 3x3, also using the corresponding AF, and 'same' padding. This is followed by another Batch Normalization layer, a MaxPooling2D layer with a pool size of 2x2, and a Dropout layer with a rate of 0.2.
- The model then repeats a similar structure: a Conv2D layer with 64 filters and the corresponding AF, a Batch Normalization layer, another Conv2D layer with 64 filters and the same activation function, another Batch Normalization layer, a MaxPooling2D layer (2x2), and a Dropout layer with a rate of 0.3.
- Again, a similar structure follows, with Conv2D layers having 128 filters, along with Batch Normalization, MaxPooling2D (2x2), and Dropout (rate 0.4) layers.
- The Conv2D layers are followed by a Flatten layer, a Dense layer with 128 units using the activation function under test, a Batch Normalization layer, a Dropout layer with a rate of 0.5, and a final Dense layer with 10 units and a softmax activation function.

For Histopathologic Cancer Detection Dataset the following CNN was used:
- The model incorporates five convolutional layers in total. The first layer employs a kernel size of 5x5, whereas the remaining four utilize a kernel size of 3x3.
- The number of convolutional filters in these layers increases in a progressive sequence, starting from 32 for the first layer and ending with 512 for the fifth layer.
- After each convolutional layer, a Max Pooling operation and Batch Normalization are applied.
- Dropout layers are also included after each convolutional layer. The dropout rates used for these layers gradually increase from 0.1 in the first layer to 0.5 in the fifth layer.
- The tested activation functions are incorporated after every convolutional layer.
- The model includes a Global Average Pooling layer, followed by the chosen activation function, Batch Normalization, and a Dropout layer with a rate of 0.3.
- A Dense layer with 256 units is then added. This layer is followed by the activation function, Batch Normalization, and a Dropout layer with a rate of 0.4.
- The final layer is a Dense layer with a softmax activation function. The number of neurons in this layer matches the number of output classes for each respective dataset.

All models are compiled with the Adam Optimizer.

## 2.2 The PLTanh Activation Function

The Rectified Linear Unit (ReLU) is among the most frequently employed activation functions (AFs) in contemporary neural networks. Its use between layers introduces nonlinearity, thus enabling the network to handle complex, nonlinear datasets. ReLU and its derivative are expressed in Eq. (1).

$$f(x) = \begin{cases} 0 \; \forall \; x < 0 \\ x \; \forall \; x > 0 \end{cases} \qquad (1)$$

$$\frac{dy}{dx} f(x) = \begin{cases} 0 \; \forall \; x < 0 \\ 1 \; \forall \; x > 0 \end{cases} \qquad (2)$$

Despite its widespread use and success in deep neural networks (DNNs), ReLU possesses some inherent drawbacks. First, ReLU is not continuously differentiable, which, while not detrimental, can slightly affect the training performance. This is due to the undefined gradient at x=0.

Furthermore, ReLU sets all values less than 0 to zero, a feature that can be advantageous for sparse data. However, the gradient of 0 is also 0, meaning that neurons reaching large negative values risk getting stuck at 0 - a phenomenon colloquially referred to as the 'dying ReLU' problem. Consequently, these 'dead' neurons halt the network's learning progression, leading to suboptimal performance.

Even with the careful initialization of weights to small random values, the summed input to the traditional ReLU AF remains negative, irrespective of the input values supplied to the neural network. To address these issues, variants of the ReLU function, such as the Leaky ReLU (LReLU), have been developed. These variants aim to deliver a more nonlinear output for small negative values or ease the transition from positive to small negative values, albeit without fully resolving the issue.

The LReLU is trying to solve these problems by providing a small negative gradient for negative inputs into a ReLU function. Fig. 1 and Eqs. (3) and (4) demonstrate the LReLU and its derivative.

$$f(x) = \begin{cases} x \ \forall \ x > 0 \\ ax \ \forall \ x \leq 0 \\ \text{where } \alpha = 0.01 \end{cases} \quad (3)$$

$$\frac{dy}{dx} f(x) = \begin{cases} 0.01 \ \forall \ x < 0 \\ 1 \ \forall \ x > 0 \end{cases} \quad (4)$$

Although theoretically LReLU is solving the 'dying ReLU', it is not actually proven to improve the classification performance. Indeed, in several studies the LReLU performance is the same or lower with ReLU.

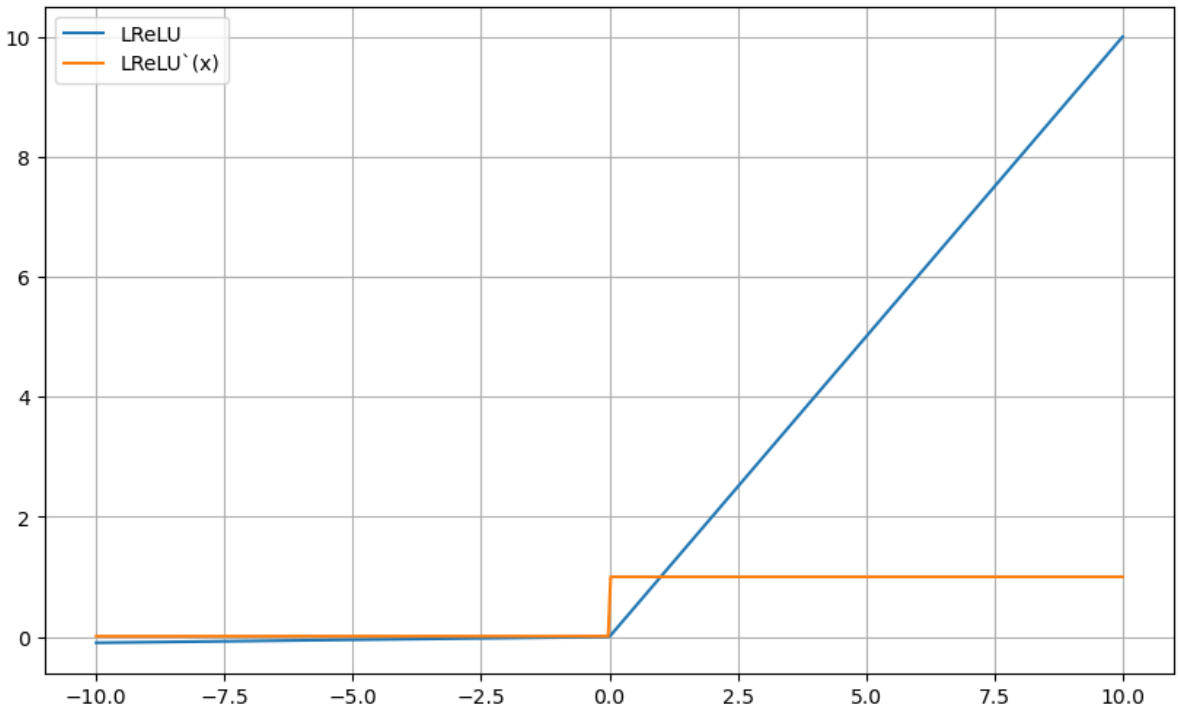

**Figure 1:** Blue: LReLU AF, Orange: LReLU Derivative

The Parametric Leaky Tanh (PLTanh) activation function introduced in this research aims to mitigate the challenges often associated with traditional Leaky ReLU and tanh functions. This activation function is given by max(tanh(x),α*abs(x)), astutely harnessing the strengths of both Tanh and Leaky ReLU.

The tanh activation function maps real numbers into the interval [-1,1], producing a normalized output that stands resilient against outliers. On the other hand, the Leaky ReLU function, apart from maintaining the positive part of its input, also introduces a slight gradient for the negative values, ensuring that all neurons remain active during the learning process. By doing so, it addresses the "dead neuron" issue, a scenario where neurons may only output zero for all inputs, often seen with the conventional ReLU. The PLTanh function synergizes the merits of these two activation functions while effectively circumventing their limitations.

The versatility of the PLTanh activation function comes from its potential to handle a diverse range of input data, thereby bolstering the learning process of deep neural networks. The inclusion of the alpha parameter offers adaptability, allowing the function to be attuned to various data distributions.

In essence, PLTanh is crafted to amalgamate the advantages of both Tanh and LReLU, while simultaneously countering their inherent weaknesses. By offering a well-adjusted response to both positive and negative inputs, it stands as a potent candidate for an efficient activation function in deep neural networks.

Fig. 3 and Eqs. (5) and (6) elucidate PLTanh and its derivative, with α=0.01, respectively

$$f(x) = \begin{cases} \tanh(x) & \forall \ \tanh(x) >= \alpha|x| \\ |\alpha|x & \forall \ \tanh(x) \le \alpha|x| \end{cases} \quad (5)$$

$$\frac{dy}{dx} f(x) = \begin{cases} 0.01 & \forall \ x>0 \land 0.01x>=\tanh(x) \\ -0.01 & \forall \ x<0 \land 0.01x+\tanh(x)<=0 \\ \text{intermediate} & \forall \ (x>=0 \land 0.01x >=\tanh(x)) \lor (x<0 \land 0.01x+\tanh(x)<=0) \\ \text{sech}^2(x) & \forall \ otherwise \end{cases}$$

(6)

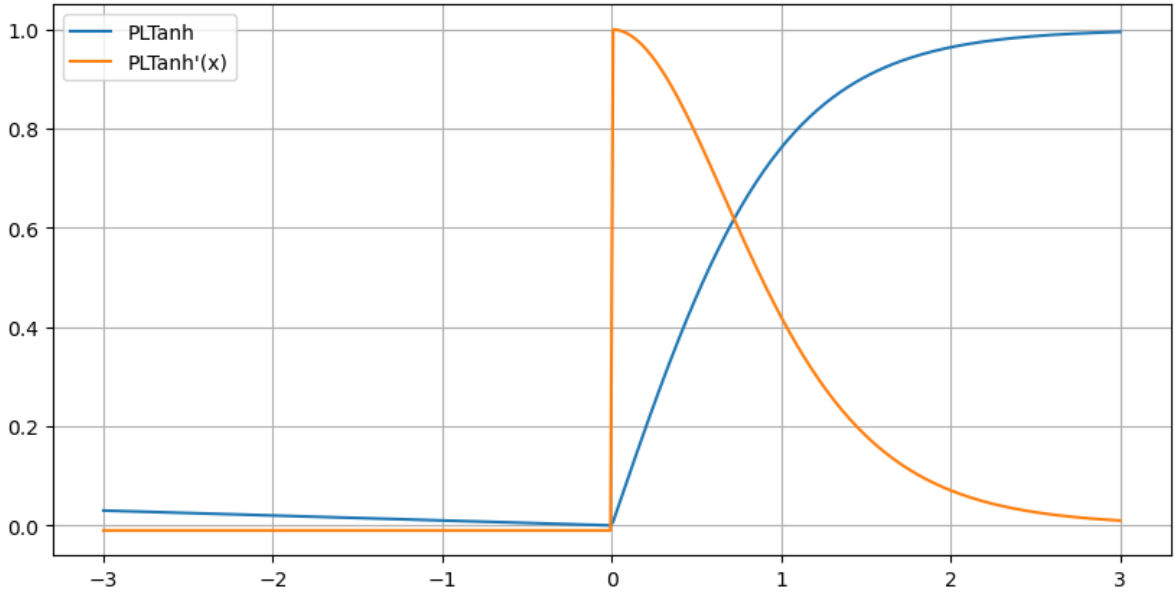

**Figure 3:** Blue: PLTanh AF, Orange: PLTanh Derivative

The derivative of the Parametric Leaky Tanh (PLTanh) function adheres to the specified rules divided into distinct conditions:

The derivative is 0.01 when x is greater than 0 and 0.01x is greater or equal to tanh(x). This reflects a 'leaky' behavior for values where x is positive or where the linear term surpasses the tanh component.

The derivative is -0.01 when x is less than 0 and 0.01x + tanh(x) is less than or equal to 0. This demonstrates the 'leaky' behavior but for negative values, effectively taking the more negative value between the linear term and the tanh component.

For intermediate scenarios, which include cases where (x is greater or equal to 0 and 0.01x is less than tanh(x)) or (x is less than 0 and 0.01x + tanh(x) is greater than 0), the structure of the derivative is not explicitly defined.

In all other circumstances, the derivative is expressed as sech^2(x), reflecting the derivative behavior of the Tanh function.

The unique configuration of the PLTanh function results in a special derivative profile. It incorporates the merits of the LReLU activation function, notably its swift learning attributes and the ability to introduce gradients for positive and slight negative inputs. Simultaneously, it taps into the Tanh activation function's ability to render a non-linear, consistently varying gradient over its entire domain. The derivative of the PLTanh function seamlessly fuses the benefits of both the LReLU and Tanh activation functions. It offers a mix of gradients – some constant, and others more dynamic – depending on the input. As such, PLTanh is adept at circumventing challenges often seen with either function in isolation, like the 'dying ReLU' phenomenon or the vanishing gradients challenge associated with the Tanh function.

Listing 1 provides the Keras implementation code for PLTanh.

**Listing 1:** A snippet of code in Python (Keras) with PLTanh implementation and usage

```python
from tensorflow.keras import backend as K
from tensorflow.keras.layers import Input, Conv2D, Lambda
from tensorflow.keras.utils import get_custom_objects
def PLTanh(x):
    alpha = 0.01
    return K.maximum(tf.keras.activations.tanh(x), alpha*K.abs(x))

get_custom_objects().update({'PLTanh':  tf.keras.layers.Activation(PLTanh)})

model = tf.keras.models.Sequential([
tf.keras.layers.Conv2D(32, kernel_size=(3, 3), activation='PLTanh', input_shape=(28, 28, 1)),
tf.keras.layers.MaxPooling2D(pool_size=(2, 2)),
tf.keras.layers.Flatten(),
tf.keras.layers.Dense(10, activation='softmax')
])
```

## 3. EXPERIMENTAL STUDY AND RESULTS

The evaluation of our trained neural network models was conducted on specific datasets using a 5-Fold cross-validation approach. This statistical method plays a crucial role in preventing overfitting, while also serving as a reliable means for comparing different learning algorithms. Moreover, it is utilized Bayesian Optimization to pinpoint the 'α' parameter for the PLTanh AF in each dataset. To ensure consistency and dependability in our results, all tests were performed on an RTX3090 GPU.

The results discussed in this section are average measures, derived from the 5-Fold cross-validation process.

| | Performance measures | PLTanh (this study) | ALReLU | LReLU | ReLU |
|---|---|---|---|---|---|
| MNIST (α = 0.01) | Macro Precision | 98.16% | 98.11% | 98.13% | 98.09% |
| | Accuracy | 98.17% | 98.11% | 98.14% | 98.09% |
| | Macro Recall | 98.16% | 98.09% | 98.12% | 98.07% |
| | AUC | 99.99% | 99.99% | 99.99% | 99.99% |
| | Macro F1 | 98.15% | 98.09% | 98.12% | 98.08% |
| Fashion MNIST (α = 0.01) | Macro Precision | 90.58% | 90.31% | 90.34% | 90.39% |
| | Accuracy | 90.33% | 90.22% | 90.21% | 90.21% |
| | Macro Recall | 90.33% | 90.22% | 90.21% | 90.21% |
| | AUC | 99.26% | 99.21% | 99.21% | 99.21% |
| | Macro F1 | 90.38% | 90.22% | 90.24% | 90.22% |
| Tf Flowers (α = 0.01) | Macro Precision | 73.36% | 72.75% | 72.50% | 73.45% |
| | Accuracy | 73.21% | 72.07% | 72.31% | 72.77% |
| | Macro Recall | 72.85% | 71.52% | 71.95% | 72.07% |
| | AUC | 92.93% | 92.48% | 92.74% | 92.57% |
| | Macro F1 | 72.58% | 71.80% | 72.01% | 72.37% |
| CIFAR-14 (α = 0.4) | Macro Precision | 85.89% | 85.36% | 85.75% | 85.60% |
| | Accuracy | 85.87% | 85.16% | 85.56% | 85.5% |
| | Macro Recall | 85.87% | 85.16% | 85.56% | 85.5% |
| | AUC | 98.81% | 98.75% | 98.78% | 98.78% |
| | Macro F1 | 85.81% | 85.11% | 85.53% | 85.46% |
| Histopathologic Cancer Det. (α = 0.000000001) | Macro Precision | 87% | 88% | 89% | 89% |
| | Accuracy | 86.68% | 86.69% | 87.34% | 87.48% |
| | Macro Recall | 87% | 87% | 87% | 87% |
| | AUC | 92.78% | 95.3% | 95.45% | 95.21% |
| | Macro F1 | 87% | 87% | 87% | 87% |

**Table 1:** Classification performance measures for ALTanh, ALReLU, ReLU and LReLU, various datasets. α=PLTanh parameter

Notably, these results underline the theoretical superiority of the proposed Parametric Leaky Tanh (PLTanh) Activation Function in handling image classification tasks. The performance metrics related to classification are depicted in Table 1, and further elaborated in the subsequent sections.

The experimental results, as indicated in the table, highlight the performance of the proposed Parametric Leaky Tanh (PLTanh) activation function, ALReLU, LReLU, and ReLU on various datasets.

For the MNIST dataset, the PLTanh model exhibited superior performance metrics, with Macro Precision, Accuracy, Macro Recall, and Macro F1 scores of 98.16%, 98.17%, 98.16%, and 98.15% respectively, slightly outperforming the other activation functions.

On the Fashion MNIST dataset, the PLTanh function also prevailed with the highest Macro Precision of 90.58%, Accuracy of 90.33%, Macro Recall of 90.33%, and Macro F1 score of 90.38%, surpassing the performance of the other functions.

In the Tf Flowers dataset, the PLTanh model demonstrated a higher Accuracy of 73.21%, Macro Recall of 72.85%, and Macro F1 score of 72.58%, as compared to the other activation functions.

On the CIFAR-10 dataset, the PLTanh model outshone the others with an Accuracy, Macro Recall, and Macro F1 score of 85.87%, 85.87%, and 85.81%, respectively.

However, for the Histopathologic Cancer Detection (Kaggle) dataset, the PLTanh model's performance was slightly lower than the others, but still commendable with an Accuracy of 86.68%, Macro Precision and Macro Recall of 87%, and a Macro F1 score of 87%.

In all cases, the Area Under the Curve (AUC) scores were remarkably high for all activation functions, indicating a strong discriminative power for the positive and negative classes.

These results collectively validate the strong performance of the PLTanh activation function, demonstrating its potential in handling various types of classification tasks.

## 4. CONCLUSION

In conclusion, this study has examined the potential of a new activation function, the Parametric Leaky Tanh (PLTanh), which is a combination of LReLU and Tanh, in comparison to existing ones such as ALReLU, LReLU, and ReLU across multiple datasets. Our experiments demonstrated that PLTanh generally exhibits superior performance metrics, outshining its counterparts in most cases. This is particularly noteworthy considering that PLTanh addresses some fundamental limitations of both LReLU and Tanh functions. PLTanh proved its robustness and efficiency in diverse datasets, with different types of data, varying from images to text. However, there is still room for improvement, as seen in the Histopathologic Cancer Detection dataset, where PLTanh was slightly outperformed by the other functions. Future work could investigate further refinement of the PLTanh parameters, aiming to further improve its generalization capabilities across a wider range of tasks. This study highlights the importance of continuous exploration in the field of neural network activation functions, driving improvements in model performance and efficiency.